\documentclass[10pt,twocolumn,letterpaper]{article}

\usepackage{iccv}
\usepackage{times}
\usepackage{epsfig}
\usepackage{graphicx}
\usepackage{amsmath}
\usepackage{amssymb}

\usepackage{array}
\usepackage{threeparttable}
\usepackage{multirow}
\usepackage{booktabs}
\usepackage{array}

\DeclareMathOperator*{\argmin}{argmin}
\usepackage{color}

\newcommand{\green}[1]{\textcolor{green}{#1}}
\usepackage[breaklinks=true,bookmarks=false]{hyperref}

\iccvfinalcopy 

\def\iccvPaperID{2478} 

\ificcvfinal\fi

\begin{document}

\title{
Customizing Student Networks From Heterogeneous Teachers \\
via Adaptive Knowledge Amalgamation}

\author{Chengchao Shen\textsuperscript{1,*}, Mengqi Xue\textsuperscript{1,*}, Xinchao Wang\textsuperscript{2}, 
Jie Song\textsuperscript{1,3}, Li Sun\textsuperscript{1}, Mingli Song\textsuperscript{1,3}\\
\textsuperscript{1}Zhejiang University, \textsuperscript{2}Stevens Institute of Technology,\\
\textsuperscript{3}Alibaba-Zhejiang University Joint Institute of Frontier Technologies \\
{\tt\small \{chengchaoshen,mqxue,sjie,lsun,brooksong\}@zju.edu.cn,xinchao.w@gmail.com}
}

\maketitle
\ificcvfinal\thispagestyle{empty}\fi

\begin{abstract}
A massive number of well-trained deep networks 
have been released by developers online. 
These networks may focus on different tasks and 
in many cases are optimized for different datasets. 
In this paper, we study how to exploit such heterogeneous pre-trained networks, 
known as teachers,
so as to train a customized student network that tackles
a set of selective tasks defined by the user. 
We assume no human annotations are available, 
and each teacher may be either single- or multi-task. 
To this end, we introduce a dual-step strategy that first extracts 
the task-specific knowledge from the heterogeneous teachers sharing
the same sub-task, and then amalgamates the extracted knowledge 
to build the student network.
To facilitate the training, 
we employ a selective learning scheme where, 
for each unlabelled sample,
the student learns adaptively from only 
the teacher with the least prediction ambiguity.
We evaluate the proposed approach on several datasets and 
experimental results demonstrate that the student, 
learned by such adaptive knowledge amalgamation, 
achieves performances even better than those of the teachers.

\end{abstract}

\renewcommand{\thefootnote}{*}
\footnotetext{Equal contribution}

\section{Introduction}
Deep networks have been applied to almost all computer vision tasks
and have achieved state-of-the-art performances, such as image classification~\cite{krizhevsky2012imagenet,simonyan2014very,he2016deep,huang2017densely}, semantic segmentation~\cite{long2015fully,chen2018deeplab,badrinarayananK17} and object detection~\cite{ren2015faster,liu2016ssd,zhang2018single}. 
This tremendous success is in part attributed to the large amount of human annotations utilized to train the parameters of the deep networks. 
In many cases, however, such training annotations are unavailable 
to the public due to for example privacy reasons. 
To reduce the re-training effort and enable the plug-and-play reproduction, 
many researchers have therefore shared online their pre-trained networks, 
which focus on different tasks or datasets.

    \begin{figure}[t]
    \centering
    \includegraphics[width=0.99\linewidth]{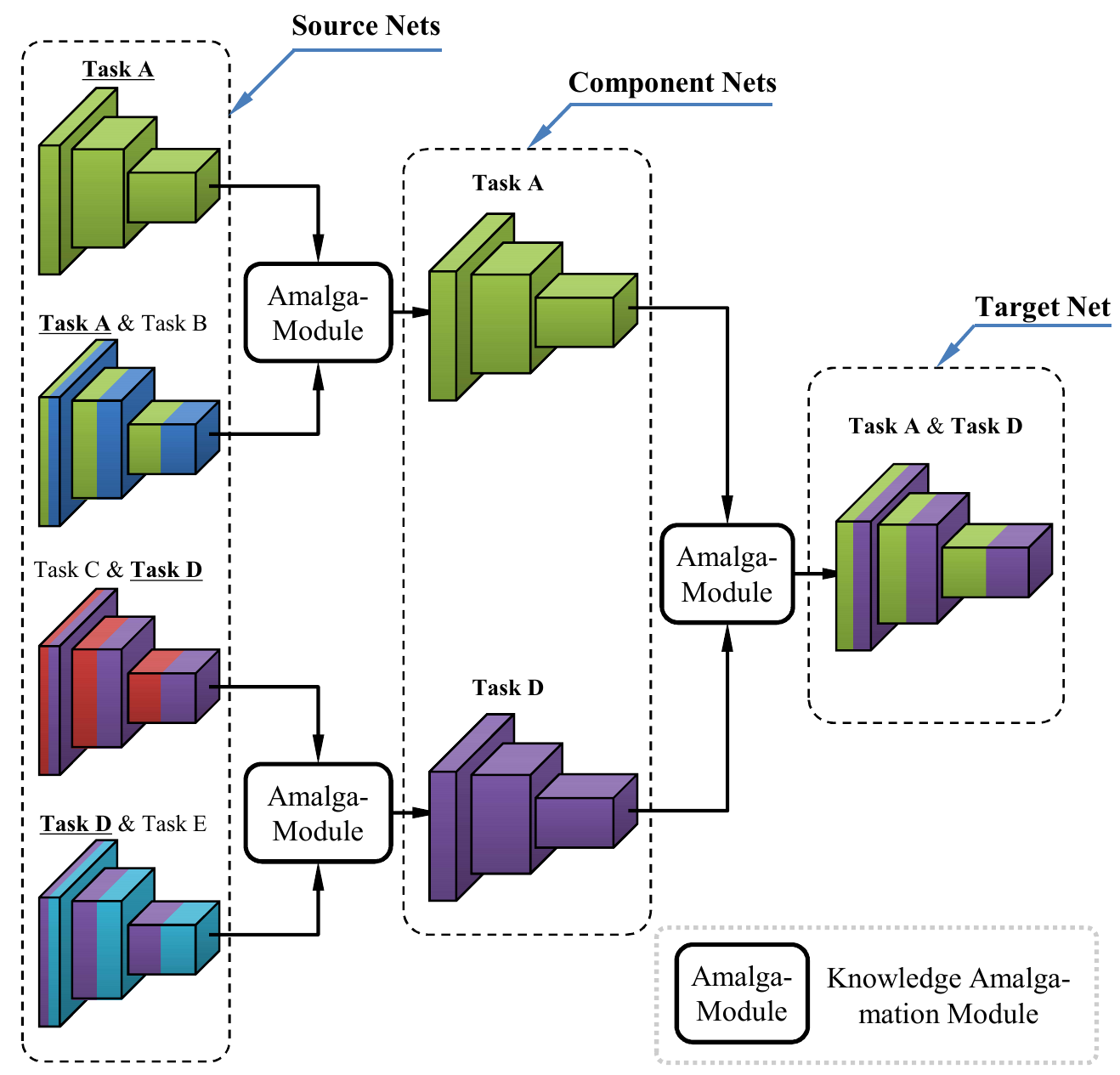}
    \caption{
    The dual-stage knowledge amalgamation strategy for customizing student networks.
    Given four source nets working on heterogeneous tasks, 
    each of which may be either single- or multi-task, 
    we cluster them into two groups, one for Task~A and the other for Task~D.
    We then conduct the first-round knowledge amalgamation for each group
    to derive the two components nets, based on which the second-round amalgamation is further carried out to produce the final student model specified by the user.
    }
    \label{fig:overview}
    \end{figure}

In this paper, we investigate how to utilize such pre-trained 
networks that focus on different tasks, 
which we term as \emph{heterogeneous teachers},
to learn a customized and multitalented network, 
termed as the \emph{student}.
We assume that we are given a pool of well-trained teachers,
yet have no access to any human annotation;
each teacher can be either single- or multi-task, 
and may or may not overlap in tasks.
Our goal is to train a compact and versatile student network
that tackles a set of selective tasks  defined by the user,
via learning from the heterogeneous teachers.
In other words, the student is expected to \emph{amalgamate}  the multidisciplinary knowledge 
scattered among the heterogeneous teachers into its compact-sized model, 
so that it is able to perform the user-specified tasks.

The merit of this customized-knowledge amalgamation problem, 
therefore, lies in that it allows for reusing pre-trained deep networks 
to build a tailored student model on user's demand, again without having to 
access human annotations. To this end, we introduce a 
dual-stage strategy that conducts knowledge amalgamation twice.
In this first stage, from the pool we pick heterogeneous 
teachers covering one or multiple desired tasks,
which we term as \emph{source nets};
we then cluster the source nets sharing the same task 
into groups, from each of which we learn a single-task network,
termed as \emph{component net}.
In the second stage, we construct our final student network, 
termed as \emph{target net}, by amalgamating the heterogeneous 
knowledge from the learned component nets.

\begin{figure*}[t]
   \centering
   \includegraphics[width=\linewidth]{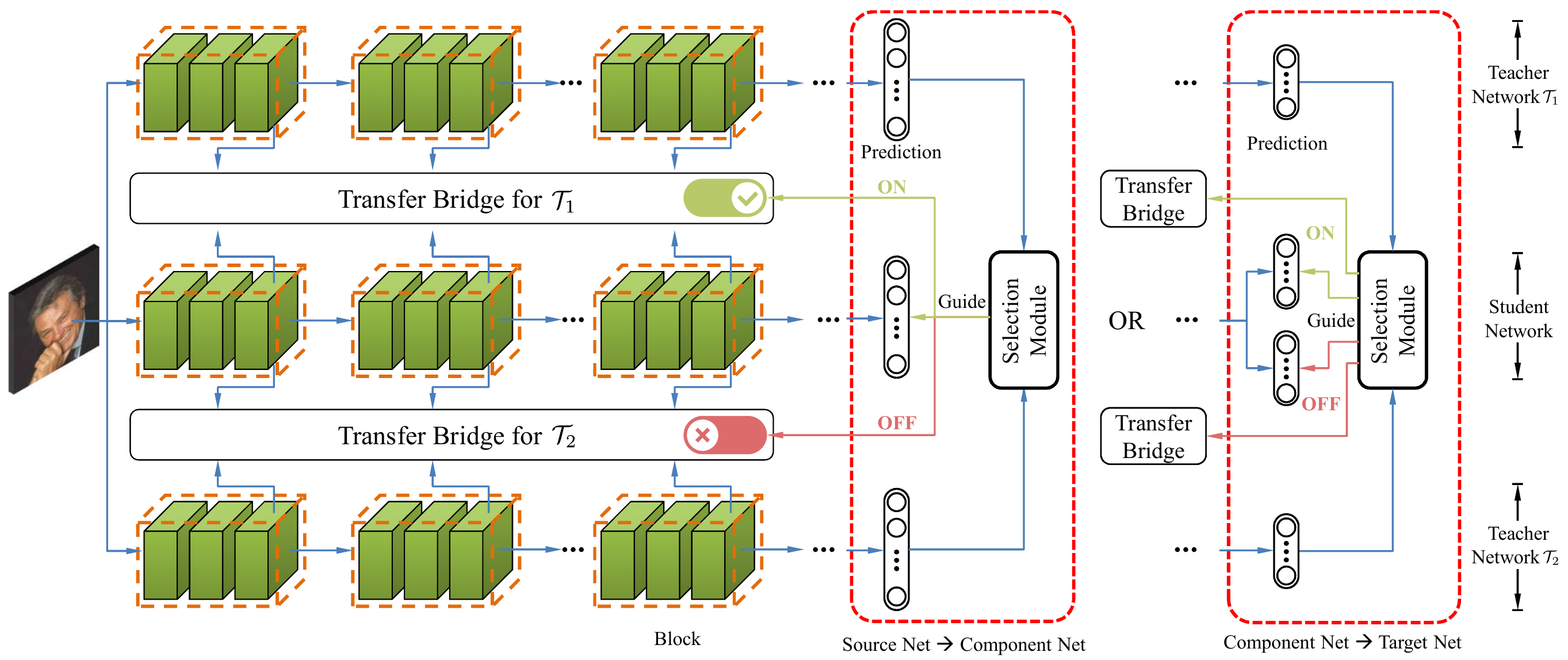}
   \caption{{
   Amalgamating knowledge from multiple teachers.
   The student learns both the predictions and the features
   from a teacher model, chosen among multiple
   via a selective learning module.
   The features of this selected teacher network are then
   transferred to the student network via 
   the transfer bridge in a block-wise manner.
   The two amalgamation steps, i.e., source-to-component and component-to-target,
   undergo the same process. }
   }
   \label{fig:knowledge-amalgamation}
   \vspace{-4mm}
\end{figure*}

We show an example  in Fig.~\ref{fig:overview} 
to illustrate our problem setup and the overall workflow.
Here we are given a pool of four source nets,
of which one is single-task and the others are multi-task.
We aim to train a compact target net, without human-labelled annotations, 
which in this case handles simultaneously Tasks~A~and~D demanded by the user.
In the first stage, we cluster the four source nets into two groups, 
one on Task~A and the other on Task~D, and learn a component net for
each task; in the second stage, we amalgamate the two component nets
to build the user-specified multi-task target net.

This dual-stage approach for knowledge amalgamation, 
as will be demonstrated in our experiments, 
turns out to be more effective than the one-shot approach
that learns a multi-task target net directly 
from  the heterogeneous source nets. 
Furthermore, it delivers the component nets as byproducts, 
which serve as the modular units that can be further integrated 
to produce any combined-task target nets, significantly 
enhancing the flexibility and modularity of the knowledge amalgamation. 

As we assume no human-labelled ground truths are provided,
it is crucial to decide which teacher among the multiple to use, 
so as to train the student effectively in both stages. 
In this regard, we exploit a selective learning scheme, where
we feed unlabelled samples to the multiple 
teacher candidates 
and allow the student to, for each sample, learn adaptively only from 
the teacher with the least prediction ambiguity.
Specifically, we adopt the chosen teacher's feature maps and 
score vectors as supervisions to train the student,
where the feature learning is achieved via 
a dedicated \emph{transfer bridge} that aligns the 
features from the teacher and the student.
Please note that, for each sample, we conduct the
teacher selection and update which teacher to learn from.

In short, our contribution is a novel knowledge amalgamation strategy
that customizes a multitalented student network 
from a pool of single- or multi-task 
teachers handling different tasks, without accessing human annotations.
This is achieved via a dual-step approach, 
where the modular and single-task component 
networks are derived in the first step
followed by their being amalgamated in the second. 
Specifically, for each unlabelled sample, 
we utilize a selective learning strategy to decide
which teacher to imitate adaptively, and introduce 
a dedicated transfer bridge for feature learning.
Experimental results on several datasets demonstrate that
the learned student models, despite their compact sizes, 
consistently outperform the teachers in their specializations.

\section{Related Work}
{
    Knowledge distillation~\cite{hinton2015distilling} 
    adopts a teacher-guiding-student strategy where a small student network learns to imitate 
    the output of a large teacher network.
    In this way, the large teacher network can transfer knowledge to the student network with smaller
    model size, which is widely applied to model compression. 
    Following~\cite{hinton2015distilling}, some works are proposed to exploit the intermediate representation
    to optimize the learning of student network, such as FitNet~\cite{romero2015fitnets}, DK$^2$PNet~\cite{wang2016accelerating},
    AT~\cite{Zagoruyko2017AT} and NST~\cite{NST2017}.
    In summary, these works pay more attention on knowledge transfer among the same classification task.
    Transfer learning is proposed to transfer knowledge from source domain to target domain to save
    data on target domain~\cite{pan2010survey}.
    It contains two main research directions: cross-domain transfer learning~\cite{long2013transfer,huang2018domain,hu2015deep,ding2018graph} 
    and cross-task one~\cite{hong2016learning,cui2018large,gholami2017punda,You2017LMT}.
    In the case of cross-domain transfer learning, the dataset adopted by source domain and the counterpart 
    of target domain are different in domain but the same in category.
    Also, cross-task transfer learning adopts the datasets that have the same domain but different categories.
    Transfer learning mainly focuses on compensating for the deficit of data on target domain with enough data
    on source domain.
    By contrast, our approach amalgamates multiple pre-trained 
    models to obtain a multitalented model using unlabelled data.

    To exploit knowledge of massive trained deep-learning-based models, researchers have made some promising attempts.
    MTZ~\cite{NIPS2018_7841} merges multiple correlated trained models by sharing neurons among these models for
    cross-model compression.
    Knowledge flow~\cite{iou2019knowledgeflow} transfers knowledge from multiple teacher models to student one with
    strategy that student learns to predict with the help of teachers, but gradually reduce the dependency on teachers,
    finally predict independently.
    Despite very promising solutions, the above approaches still depend on labelled dataset, 
    which is not suitable for our application scenario where no human labels are available.
    
    The approach of~\cite{shen2019amalgamating} proposes to
    transfer knowledge from multiple trained models into a single one in a layer-wise
    manner with unlabelled dataset.
    It adopts an auto-encoder architecture to amalgamate features from multiple 
    single-task teachers. 
    Several knowledge amalgamation methods are also proposed to handle the above 
    task~\cite{ye2019student,luo2019knowledge,Ye2019Amalgamating}.
    The proposed approach here, on the other hand,
    handles teachers working on both single or multiple tasks, and
    follows a dual-stage strategy tailored for customizing the student network
    that also gives rise to component nets as byproducts.

}


\section{The Proposed Approach}

In this section, we give more details on the proposed approach for 
customizing multi-task students. We first give an overview of the overall process,
then introduce the {transfer bridge} for learning the features of the student, afterwards we describe the selective learning scheme for  choosing teachers adaptively, and finally show the loss function.

\subsection{Overview}
{
    Our problem setup is as follows. We assume that we 
    are given a pool of pre-trained source nets,
    each of which may be single- or multi-task, 
    where the former can be treated as a degenerated case of the latter. 
    These source nets may be trained for distinct tasks and optimized for different datasets.
    Let $K_i$ denote the set of tasks handled by source net~$i$, and let 
    $\mathcal{K} =  \bigcup_i K_i $ denote the set of tasks covered by all the teachers.
    Our goal is to customize a student model that tackles 
    a set of user-specified tasks, denoted by $K_s \subseteq \mathcal{K}$. Also,
    we use $M_s$ to denote the number of tasks to be trained for the student, 
    i.e.,  $|K_s| = M_s$. As the initial attempt along this line, 
    for now we focus on image classification and assume the source nets all take the form of 
    the widely-adopted resnet~\cite{he2016deep}. The proposed approach,
    however, is not restricted to resnet and is applicable to other architectures as well. 
    
    To this end, we adopt a dual-stage strategy 
    to conduct the selective knowledge amalgamation. 
    In the first stage, we pick all the source nets that cover one or multiple tasks
    specified by the users, i.e., $i: K_i \cap K_s  \neq \varnothing$,
    and then cluster them into $M_s$ groups, each of which focus on one task only.
    For each such group we carry out the first-round knowledge amalgamation and derive
    a component net tailored for each task, all of which together 
    are further amalgamated again in the second round 
    to form the final multi-task target network.
    
    The two rounds of knowledge amalgamation are achieved 
    in a similar manner, as depicted in Fig.~\ref{fig:knowledge-amalgamation}.
    In the first round, we refer to the source and the component respectively as teachers and students, and 
    in the second, we refer to the component and the target respectively as teachers and student. 
    Specifically, we conduct a block-wise learning scheme, 
    as also done in~\cite{szegedy2015going,he2016deep,huang2017densely},
    where a transfer bridge is established between each teacher and the student
    so as to allow the student to imitate the features of the teacher. 
    {In both amalgamation rounds,  
    for each unlabelled sample,
    student adaptively learns from only
    one selected teacher,
    which is taken to be the one 
    that yields the least prediction ambiguity.
    For }
    In what follows, we introduce the proposed transfer bridge and the selective learning strategy in details.
    
}

\subsection{Transfer Bridge}\label{sec:transfer_bridge}
{
    A transfer bridge, as shown in Fig.~\ref{fig:transfer_bridge},
    is set up between the student and each teacher, 
    in aim to align the features of the student and the teachers 
    so that the former can learn from the latter. 
    As the teachers may be multi-task and therefore
    comprise knowledge not at the interest of the student, 
    we would have to ``filter'' and transform
    the related features from the teacher
    in a way that is learnable by the student. 
    This is achieved via a dedicated feature alignment~(FA) module
    and a regularized loss function, discussed as follows.

\begin{figure}[t]
        \centering
        \includegraphics[width=1.02\linewidth]{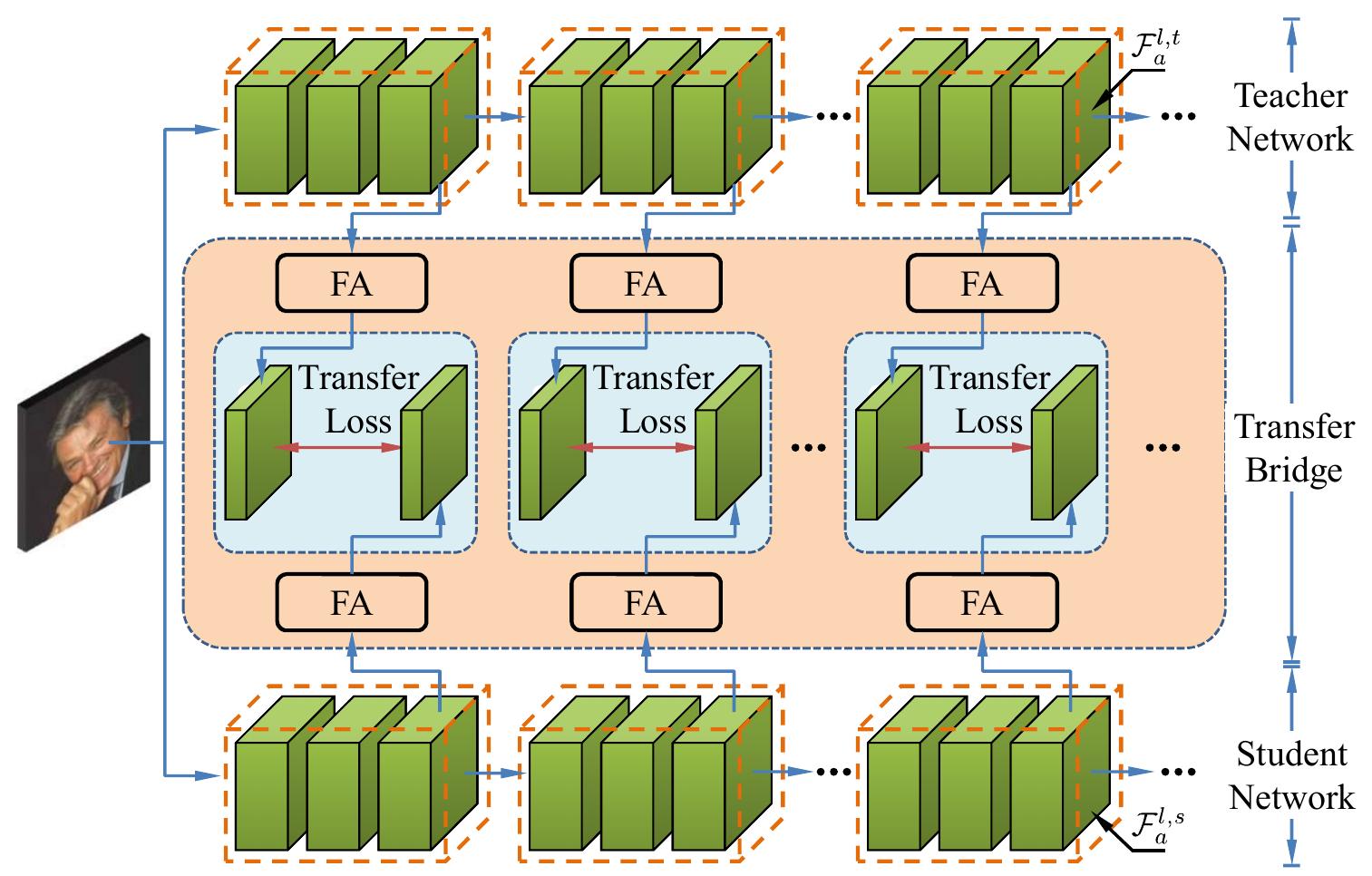}
        \caption{Transfer bridge between a teacher network and a student.
        The block-wise features from teacher and student are respectively transformed into $\mathcal{F}_{a}^t$ 
        and $\mathcal{F}_{a}^s$ by the FA module, which
        are then utilized for computing the transfer loss.
        }
        \label{fig:transfer_bridge}
    \end{figure}

    {\bf Feature Alignment~(FA).}
    An FA module, which learns to filter and align the target-task-related features,
    is introduced between each block of the teacher and the student.
    In our implementation, FA takes the form of 
    an $1\times1$ convolutional operation~\cite{shen2019amalgamating,szegedy2015going,Li2018FewSK}.
    As depicted in Fig.~\ref{fig:fsa}, the feature maps of both the student and the teacher
    are weighted and summed to obtain a new feature map across channels
    by the $1\times1$ convolutional operation. We write,
    \begin{align} \label{eq:encode}
        \mathcal{F}_{a,c} = \sum\limits_{c'=1}^{C_{in}} w_{c, c'} \mathcal{F}_{c'},
    \end{align}
    where $\mathcal{F}_{a,c}$ denotes the $c$-th channel of aligned feature maps $\mathcal{F}_{a}$, 
    $\mathcal{F}_{c'}$ denotes the $c'$-th channel of input feature maps from the teacher or the student, and $w_{c, c'}$ denotes the weight of $1\times1$ convolutional operation, which transforms  $\mathcal{F}_{c'}$ to $\mathcal{F}_{a,c}$.

     \begin{figure}[t]
        \centering
        \includegraphics[width=\linewidth]{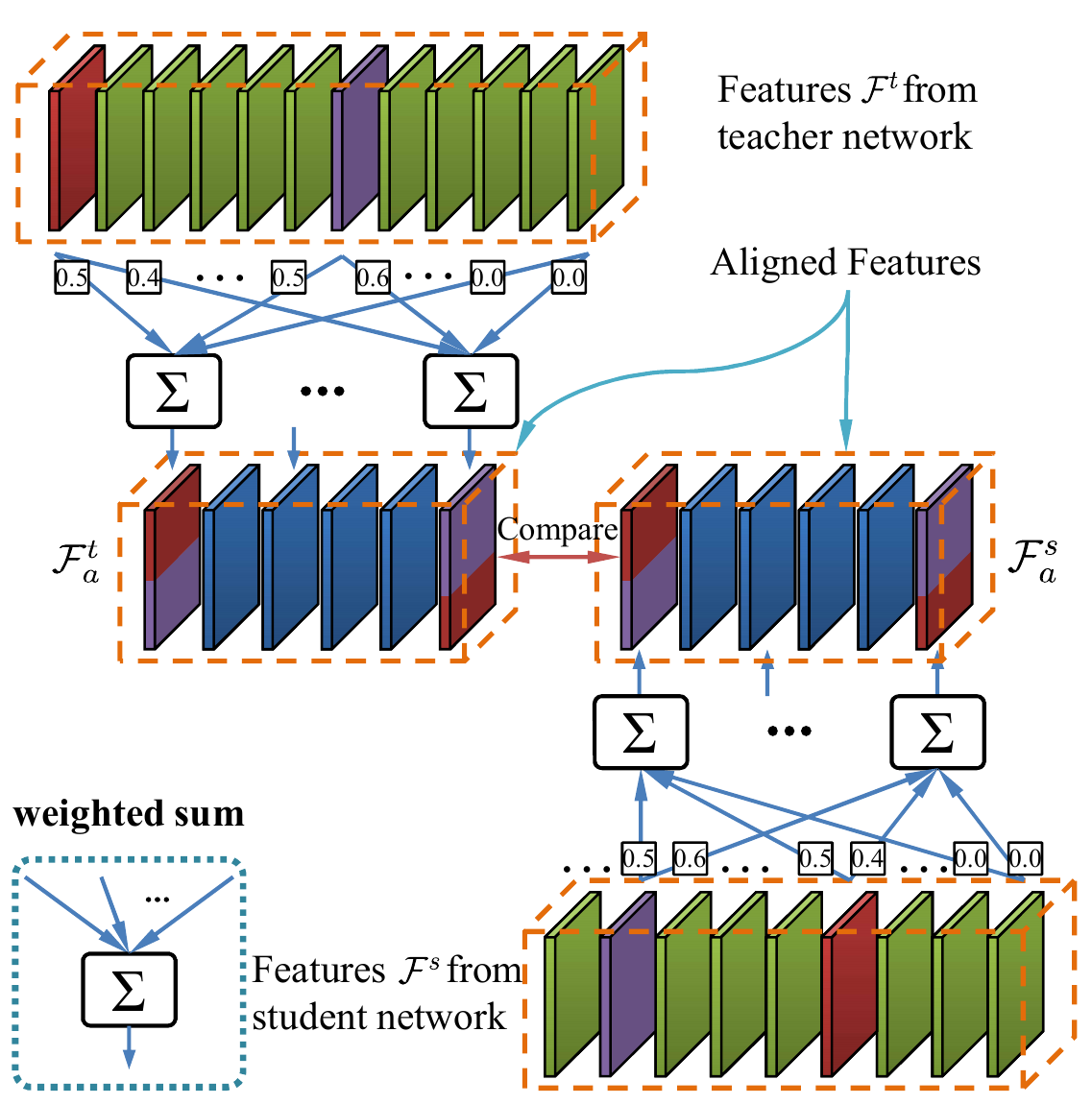}
        \caption{Feature alignment.
        The features from teacher network and student network are 
        transformed and aligned using $1 \times 1$
        convolutional operation. 
        }
        \label{fig:fsa}
    \end{figure}

    {\bf Transfer Loss and Weight Regularization. }
    To supervise the feature learning, we define a transfer loss
    based on the aligned features of the teacher and the student.
    Let $\mathcal{F}_{a}^{l, t}$ denote the feature maps from block $l$ 
    of the teacher network  and let $\mathcal{F}_{a}^{l,s}$
    denote those of the student. 
    We first introduce the vanilla transfer loss, as follows,
    \begin{align} \label{eq:transfer_loss}
        \mathcal{L}_{a}^{l, t}  =  \frac{1}{C_{out}^l H^l W^l} \Vert \mathcal{F}_{a}^{l,s} - \mathcal{F}_{a}^{l, t} \Vert^2,
    \end{align}
    where $C_{out}^l$, $H^l$ and $W^l$ denotes the channel, 
    height and width size of $\mathcal{F}_{a}^{l, t}$ or
    $\mathcal{F}_{a}^{l, s}$, respectively.

    This vanilla transfer loss alone, however, may lead to
    trivial solutions: by taking the two features maps to be zero,
    the loss collapses to zero. To avoid such degenerated case, 
    we impose a regularization on the transfer loss. 
    As the aligned features are controlled by the learnable parameters
    $w_{ij}$, we introduce a constraint of $w_{ij}$, 
    as follows,
    \begin{align} \label{eq:constraint}
        \sum \limits_{i=1}^{C_{in}^l} w_{ij}^{(l, t)2} =1,
    \end{align}
    which on the one hand limits the magnitude of $w_{ij}$
    to a reasonable range, and on the other hand eliminate the trivial solutions.
    For the sake of optimization, we then relax the above hard constraint into
    a soft one:
    \begin{align} \label{eq:regularization}
        \mathcal{L}_{reg}^{l,t}=\frac{1}{C_{out}^l} \sum \limits_{j=1}^{C_{out}^l}(\sum \limits_{i=1}^{C_{in}^l} w_{ij}^{(l,t)2} - 1)^2,
    \end{align}
    which are further added to the final loss described in Sec.~\ref{sec:loss}.

}

\subsection{Selective Learning}
{
    As we assume no ground-truth annotations are given for training the student
    and meanwhile multiple teachers handling the same task might be available,
    we would have to ensure that, for each unlabelled sample, 
    we allow the student to learn from only the ``best'' possible teacher among many. 
    Since there are, again, no 
    ground truths for evaluating the teacher with
    the best sample-wise performance, we resort to  
    learning from the teacher with the most ``confident''
    prediction. In other words, the student imitates
    the predictions and features of the teacher with the least
    prediction ambiguity. 
    
    Here, we use entropy impurity to measure the prediction ambiguity:
    the smaller the value is, the higher the confidence of prediction is.
    The teacher with minimal entropy impurity is therefore 
    selected to guide
    the learning of student network:
    \begin{align} \label{eq:entropy_metric}
        {\rm I}(p^t(x)) =  -\sum \limits_{i} p_i^t(x)\log(p_i^t(x)), \\
        t_{se} = \argmin \limits_{t} {\rm I}(p^t(x)),
    \end{align}
    where $t_{se}$ indexes the selected teacher.
    
}

\subsection{Loss Function}
\label{sec:loss}
{
    {
    To imitate the predictions of teachers, 
    we introduce a soft target loss 
    between the predictions of teacher networks and that of the student.
    Since the student is required to learn multiple teachers 
    and the outputs of teachers are typically 
    different from each other, 
    a learnable scale parameter $\lambda_t$ is introduced to compensate such scale
    difference. We write,
    \begin{align} \label{eq:scale}
        \mathcal{L}_{soft}^{t} = \frac{1}{C_{cls}}\Vert \mathcal{F}_{score}^{s} - \lambda_t \mathcal{F}_{score}^{t} \Vert^2, 
    \end{align}
    where $\mathcal{F}_{score}^{s}$ and $\mathcal{F}_{score}^{t}$ 
    denote the logits before softmax layer from
    student and teacher, respectively, and $C_{cls}$ denotes the length of logits.}
    
    {The total loss of knowledge amalgamation 
    between source nets and component net and the one between component nets and target
    net are defined as follows,}
    \begin{align} \label{eq:total_loss_s2c}
        \mathcal{L}_{total} = \sum \limits_{l=1}^{L-1} \{\mathcal{L}_a^{(l,t_{se})} + \mathcal{L}_{reg}^{(l,t_{se})}\} + \mathcal{L}_{soft}^{t_{se}}, 
    \end{align}
    where $L$ denotes the number of blocks in source, component or target net.

}

\section{Experiments}
{
    In this section, we show the experimental results of the proposed
    adaptive knowledge amalgamation. We start by introducing the datasets we used
    and the implementation details, and then provide the quantitative analysis
    including results on attribute and category classification. 
    More results and additional details can be found in the supplementary material.

    {
        \begin{table}[]
        \begin{center}
        \begin{tabular}{|c|c|c|}
        \hline
        \multirow{2}{*}{{\bf Dataset}} & \multicolumn{2}{c|}{{\bf Partition}}\\ \cline{2-3}
        & {\bf 2 parts} & {\bf 4 parts} \\ \hline
        Stanford Dogs & $\mathcal{D}_1$, $\mathcal{D}_2$ & $\mathcal{D}_1'$, $\mathcal{D}_2'$, $\mathcal{D}_3'$, $\mathcal{D}_4'$ \\ \hline
        CUB-200-2011 & $\mathcal{B}_1$, $\mathcal{B}_2$ & $\mathcal{B}_1'$, $\mathcal{B}_2'$, $\mathcal{B}_3'$, $\mathcal{B}_4'$ \\ \hline
        FGVC-Aircaft & $\mathcal{A}_1$, $\mathcal{A}_2$ & $\mathcal{A}_1'$, $\mathcal{A}_2'$, $\mathcal{A}_3'$, $\mathcal{A}_4'$ \\ \hline
        Cars & $\mathcal{C}_1$, $\mathcal{C}_2$ & $\mathcal{C}_1'$, $\mathcal{C}_2'$, $\mathcal{C}_3'$, $\mathcal{C}_4'$ \\ \hline
        \end{tabular}
        \end{center}
        \caption{The partition of four fine-grained datasets for the training of source nets. Each set contains the same number of categories.}
        \label{table:partion}
        \end{table} 
    }

    {
        \begin{table*}[]
        \begin{center}
        \begin{tabular}{|c|m{6cm}|c|m{6cm}|}
        \hline
        \textbf{Source Net} & \multicolumn{1}{c|}{\textbf{Attributes}} & \textbf{Source Net} & \multicolumn{1}{c|}{\textbf{Attributes}} \\ \hline
        $\mathcal{S}_1^\text{\tiny mouth}$ & big lips, narrow eyes, pale skin & $\mathcal{S}_6^\text{\tiny mouth}$ & mouth slightly open \\ \hline
        $\mathcal{S}_2^\text{\tiny mouth}$ & big lips, chubby, young & $\mathcal{S}_7^\text{\tiny mouth}$ & mouth slightly open, chubby blurry, blond hair \\ \hline
        $\mathcal{S}_3^\text{\tiny mouth}$ & smiling, arched eyebrows, attractive, black hair & $\mathcal{S}_8^\text{\tiny mouth}$ & wearing lipstick, arched eyebrows, attractive \\ \hline
        $\mathcal{S}_4^\text{\tiny mouth}$ & smiling, bags under eyes, blurry, blond hair & $\mathcal{S}_9^\text{\tiny mouth}$ & wearing lipstick, bags under eyes, blurry \\ \hline
        $\mathcal{S}_5^\text{\tiny mouth}$ & smiling, bushy eyebrows, oval face, brown hair & $\mathcal{S}_{10}^\text{\tiny mouth}$ & wearing lipstick, bushy eyebrows, oval face \\ \hline
        \end{tabular}
        \end{center}
        \caption{Source nets that work on multiple attribute recognition tasks on the CelebA dataset.}
        \label{table:source-net}
        \end{table*}

        \begin{table*}[]
        \begin{center}
        \begin{threeparttable}
        \begin{tabular}{p{2.5cm}<{\centering}|p{3.1cm}<{\centering}|p{3.2cm}<{\centering}|p{3.1cm}<{\centering}|p{3.2cm}<{\centering}}
        \hline
        \multirow{2}{*}{\textbf{Model}} & \multicolumn{4}{c}{\textbf{Mouth-Related Attributes}}\\ \cline{2-5}
         & Big Lips & Smiling & Mouth Slightly Open & Wearing Lipstick \\ \hline
        Source Net & \begin{tabular}[c]{@{}c@{}}$\mathcal{S}_1^\text{\tiny mouth}$(68.7), $\mathcal{S}_2^\text{\tiny mouth}$(68.5)\end{tabular} & \begin{tabular}[c]{@{}c@{}}$\mathcal{S}_3^\text{\tiny mouth}$(88.6), $\mathcal{S}_4^\text{\tiny mouth}$(88.6), \\$\mathcal{S}_5^\text{\tiny mouth}$(87.5)\end{tabular} & \begin{tabular}[c]{@{}c@{}} $\mathcal{S}_6^\text{\tiny mouth}$(89.6), $\mathcal{S}_7^\text{\tiny mouth}$(89.5)\end{tabular} & \begin{tabular}[c]{@{}c@{}} $\mathcal{S}_8^\text{\tiny mouth}$(90.4), $\mathcal{S}_9^\text{\tiny mouth}$(90.4), \\ $\mathcal{S}_{10}^\text{\tiny mouth}$(90.3)\end{tabular} \\ \hline
        Component Net & $69.2^{{\bf \green{\uparrow 0.5,0.7}}}$ & $90.5^{{\bf \green{\uparrow 1.9,1.9,3.0}}}$ & $91.4^{{\bf \green{\uparrow 1.8,1.9}}}$ & $91.7^{{\bf \green{\uparrow 1.3,1.3,1.4}}}$ \\ \hline
        Target Net & $69.2^{{\bf \green{\uparrow 0.5,0.7}}}$ & $90.8^{{\bf \green{\uparrow 2.2,2.2,3.3}}}$ & $91.4^{{\bf \green{\uparrow 1.8,1.9}}}$ & $91.8^{{\bf \green{\uparrow 1.4,1.4,1.5}}}$ \\ \hline
        \end{tabular}
        \vspace{1mm}
        \begin{tabular}{p{2.5cm}<{\centering}|p{3.1cm}<{\centering}|p{3.2cm}<{\centering}|p{3.1cm}<{\centering}|p{3.2cm}<{\centering}}
        \hline
        \multirow{2}{*}{\textbf{Model}} & \multicolumn{4}{c}{\textbf{Hair-Related Attributes}}\\ \cline{2-5}
         & Black Hair & Blond Hair & Brown Hair & Bangs \\ \hline
        Source Net & \begin{tabular}[c]{@{}c@{}}$\mathcal{S}_1^\text{\tiny hair}$(85.2), $\mathcal{S}_2^\text{\tiny hair}$(86.9)\end{tabular} & \begin{tabular}[c]{@{}c@{}}$\mathcal{S}_3^\text{\tiny hair}$(94.0), $\mathcal{S}_4^\text{\tiny hair}$(94.2) \end{tabular} & \begin{tabular}[c]{@{}c@{}} $\mathcal{S}_5^\text{\tiny hair}$(86.4), $\mathcal{S}_6^\text{\tiny hair}$(86.3), \\ $\mathcal{S}_7^\text{\tiny hair}$(86.7)\end{tabular} & \begin{tabular}[c]{@{}c@{}} $\mathcal{S}_8^\text{\tiny hair}$(94.5), $\mathcal{S}_9^\text{\tiny hair}$(94.4)\end{tabular} \\ \hline
        Component Net & $87.8^{{\bf \green{\uparrow 2.6, 0.9}}}$ & $95.0^{{\bf \green{\uparrow 1.0, 0.8}}}$ & $88.0^{{\bf \green{\uparrow 1.6, 1.7, 1.3}}}$ & $95.2^{{\bf \green{\uparrow 0.7, 0.8}}}$ \\ \hline
        Target Net & $87.9^{{\bf \green{\uparrow 2.7, 1.0}}}$ & $95.0^{{\bf \green{\uparrow 1.0, 0.8}}}$ & $88.1^{{\bf \green{\uparrow 1.7, 1.8,1.4}}}$ & $95.2^{{\bf \green{\uparrow 0.7, 0.8}}}$ \\ \hline
        \end{tabular}
        \begin{tablenotes}
        \item[\green{$\uparrow$}] denotes performance improvement compared with the corresponding source net.
        \end{tablenotes}
        \end{threeparttable}
        \end{center}
        \caption{The performance (\%) of knowledge amalgamation from source nets to component net and
        from component nets to target net on the CelebA dataset.
        Number in parentheses denotes the accuracy of the corresponding source net.
        Unlike the component net handles only one task, the target net handles four tasks simultaneously.
        }
        \label{table:attribute}
        \end{table*}
    }

    \subsection{Experiment Settings}
    {
        
        \subsubsection{Datasets}
        {
            CelebFaces Attributes Dataset (CelebA)~\cite{liu2015faceattributes} is a large-scale face attributes 
            dataset, which consists of more than 200K celebrity images, each with 40 attribute annotations.
            It contains $162,770$ images for training, $19,868$ images for validation and $19,962$ ones for testing.
            Due to it's large size and massive attribute annotations, it can be used to build a well-trained source
            network pool to verify the proposed approach.
            We randomly split the training set into six parts with the same size, in which five parts are used to 
            train five different multi-task teachers and the remaining one is used as unlabelled training data for 
            the student.
            The experiments of network customization are conducted on two attribute groups: mouth-related attributes
            and hair-related attributes.
            More experiments on other attribute groups can be found in the supplementary material.

            Besides experiments on attribute recognition, four fine-grained datasets are used to evaluate the 
            effectiveness on network customization of category recognition.
            Stanford Dogs~\cite{Khosla_FGVC2011} contains $12,000$ images about 120 different kinds of dogs.
            FGVC-Aircraft~\cite{maji13fine-grained} consists of $10,000$ images of 100 aircraft variants.
            CUB-200-2011~\cite{WahCUB_200_2011} is a bird dataset, which includes $11,788$ images from 200 bird species.
            Cars~\cite{KrauseStarkDengFei-Fei_3DRR2013} comprises $16,185$ images of 196 classes of cars.
            The four datasets can be categorized into two groups: animal-related and vehicle-related dataset.
            As shown in Tab.~\ref{table:partion}, 
            all datasets are randomly split into several sets, 
            each of which 
            contains the same number of categories.
            For example, both $\mathcal{D}_1$ and $\mathcal{D}_2$ contain 60 breeds of dogs, $\mathcal{D}_1'$ to 
            $\mathcal{D}_4'$ contain 30 breeds of dogs, respectively.
            The details of each set can be found in the supplementary material.

        }

        \subsubsection{Implementation}
        {
            The proposed method is implemented by PyTorch on a Quadro M6000 GPU.
            The source nets adopt the same network architecture: resnet-18~\cite{he2016deep}, which are trained
            by finetuning the ImageNet pretrained model.
            Both component net and target net adopt resnet-18-like network architectures.
            The adopted net has the same net structure as the original resnet-18, except the channel number of feature maps.
            For example, the target net amalgamates knowledge from multiple component nets, so the target net
            should be more "knowledgeable" than a single component net, which should have more channels than component net.
            More implementation details can be found in the supplementary material.
        }

    }

    \subsection{Experimental Results}
    {
    In what follows, we show network customization results 
    for attribute- and category-classification, learning from
    various numbers of teachers,
    ablation studies by turning off some of the modules,
    as well as the results of one-shot amalgamation. 
    
        \subsubsection{{Network Customization for Attribute}}
        {

            In the first amalgamation step, multiple related source nets are amalgamated into a single component net
            to obtain a component task.
            Tab.~\ref{table:source-net} collects 10 source nets, 
            each of which contains a mouth-related attribute recognition
            task.
            For example, $\mathcal{S}_1^\text{\tiny mouth}$ is a source net for multiple tasks: ``big lips'', ``narrow eyes''
            and ``pale skin'', including a mouth-related attribute task: ``big lips''.
            Combined with $\mathcal{S}_2^\text{\tiny mouth}$ that also works on ``big lips'' task, 
            they are amalgamated into a component net for ``big lips'' task, as shown in Tab.~\ref{table:attribute}.
            In the second amalgamation step, multiple component nets specified by user are amalgamated into the target net.
            In Tab.~\ref{table:attribute}, the component nets about mouth-related attributes: ``big lips'', ``smiling'', 
            ``mouth slightly open'', and ``wearing lipstick'' are used to customize the corresponding target net.

            From Tab.~\ref{table:attribute}, we 
            observe consistent experimental results on two attribute 
            groups\footnote{The lookup table for the hair-related source nets as Tab.~\ref{table:source-net} is 
            provided in the supplementary material.} as follows.
            On the one hand, the performance of component net is 
            superior to those of the corresponding source nets.
            Also, the obtained component nets are more compact 
            than the ensemble of all source nets, as shown in Tab.~\ref{table:resource}.
            In particular, for ``smiling'' attribute, the component net outperforms 
            the source net $\mathcal{S}_5^\text{\tiny mouth}$ by 3.0\%.
            It supports that our approach is indeed able to
            transfer knowledge from multiple source nets into the component net, 
            and the transferred knowledge can significantly supplement the knowledge deficiency of a single source net.
            On the other hand, the target net achieves 
            comparable or better performance on the corresponding tasks,
            yet is more resource-efficient. The net parameters and computation load 
            (FLOPs: Float Operations) of target net,
            as shown in Tab.~\ref{table:resource},
            are much lower than the summation of all component nets,

            To validate the flexibility of network customization, 
            we also customize target net with different numbers of 
            component nets, for which the results are 
            shown in Tab.~\ref{table:flexible}.
            These results demonstrate that our proposed approach can be competent to the customization for different numbers
            of component nets.

            {
                \begin{table}[]
                \begin{center}
                \begin{tabular}{ccc}
                \toprule
                Model & Parameters & FLOPs \\ 
                \midrule
                Source Nets & 111.8M & 36.3G \\ 
                Component Nets & 44.8M & 14.5G \\ 
                Target Net & 22.1M & 7.0G \\ 
                \bottomrule
                \end{tabular}
                \end{center}
                \caption{The comparison of resource required in 10 source nets, 4 component nets and target net in Tab.~\ref{table:attribute}, 
                including the number of parameters and FLOPs.}
                \label{table:resource}
                \end{table} 
            }

            {
                \begin{table}[]
                \begin{center}
                \begin{tabular}{|c|p{16mm}<{\centering}|p{16mm}<{\centering}|p{16mm}<{\centering}|}
                \hline
                \multirow{5}{*}{{\bf Model}} & \multicolumn{3}{c|}{\bf Mouth-Related Attributes} \\ \cline{2-4}
                & \begin{tabular}[c]{@{}c@{}}smiling\\ lipstick \end{tabular} & 
                \begin{tabular}[c]{@{}c@{}}smiling \\ mouth open \\ lipstick\end{tabular} & 
                \begin{tabular}[c]{@{}c@{}}big lips\\ smiling \\ mouth open \\ lipstick\end{tabular} \\ \hline
                Target Net & 
                \begin{tabular}[c]{@{}c@{}}91.1 \\ 91.9\end{tabular} & 
                \begin{tabular}[c]{@{}c@{}}91.2\\ 91.7 \\ 91.7\end{tabular} & 
                \begin{tabular}[c]{@{}c@{}}69.2\\ 90.8\\ 91.4 \\ 91.8\end{tabular} \\ \hline
                \end{tabular}

                \vspace{1mm}
                \begin{tabular}{|c|p{16mm}<{\centering}|p{16mm}<{\centering}|p{16mm}<{\centering}|}
                \hline
                \multirow{5}{*}{{\bf Model}} & \multicolumn{3}{c|}{\bf Hair-Related Attributes} \\ \cline{2-4}
                & \begin{tabular}[c]{@{}c@{}}black hair\\ brown hair\end{tabular} & 
                \begin{tabular}[c]{@{}c@{}}black hair\\ brown hair\\ bangs\end{tabular} & 
                \begin{tabular}[c]{@{}c@{}}black hair\\ blond hair\\ brown hair \\ bangs\end{tabular} \\ \hline
                Target Net & 
                \begin{tabular}[c]{@{}c@{}}87.8\\ 88.2 \end{tabular} & 
                \begin{tabular}[c]{@{}c@{}}87.7\\ 88.1\\ 95.2\end{tabular} & 
                \begin{tabular}[c]{@{}c@{}}87.9\\ 95.0\\ 88.1 \\ 95.2\end{tabular} \\ \hline
                \end{tabular}
                \end{center}
                \caption{The performance (\%) of the customization of target net with different numbers of component nets on the CelebA dataset.}
                \label{table:flexible}
                \vspace{-3mm}
                \end{table}
            }
        }

        \subsubsection{{Network Customization for Category}}
        {
            We also conduct network customization experiments on category recognition.
            As shown in Tab.~\ref{table:srcnet-category}, source nets on four datasets are provided.
            For example, source net for part of Stanford Dogs $\mathcal{D}_1$: $\mathcal{S}_1^\text{\tiny dog}$ is trained 
            on the category sets $\mathcal{D}_1$ and $\mathcal{B}_1'$.
            The source nets for Stanford Dogs include $\mathcal{S}_1^\text{\tiny dog}$ and $\mathcal{S}_2^\text{\tiny dog}$
            for $\mathcal{D}_1$, $\mathcal{S}_3^\text{\tiny dog}$ and $\mathcal{S}_4^\text{\tiny dog}$ for $\mathcal{D}_2$.
            To customize a target net for category set $\mathcal{D}_1 \cup \mathcal{D}_2$, 
            the dual-step amalgamation is 
            implemented as follows.
            In the first step, source nets $\mathcal{S}_1^\text{\tiny dog}$ and $\mathcal{S}_2^\text{\tiny dog}$ are 
            amalgamated into a component net for $\mathcal{D}_1$.
            In the same way, source nets $\mathcal{S}_3^\text{\tiny dog}$ and $\mathcal{S}_4^\text{\tiny dog}$ are 
            amalgamated into a component net for $\mathcal{D}_2$.
            In the second step, component nets for $\mathcal{D}_1$ and $\mathcal{D}_2$ are amalgamated into the final target net.
            Experiments on the remaining datasets are implemented in the same way.

            The experimental results shown in Tab.~\ref{table:category} demonstrate that the component nets consistently
            outperform the corresponding source nets, and the target net achieves comparable or better accuracy than the 
            corresponding component net.
            It supports that our proposed method also works on category recognition task.

            {
                \begin{table}[]
                \begin{center}
                \begin{tabular}{|c|c|c|c|c|}
                \hline
                \multirow{2}{*}{{\bf Dataset}} & \multicolumn{4}{c|}{{\bf Source Nets}} \\ \cline{2-5}
                & $\mathcal{S}_1$ & $\mathcal{S}_2$ & $\mathcal{S}_3$ & $\mathcal{S}_4$ \\ \hline
                Dogs & \begin{tabular}[c]{@{}p{4mm}<{\centering}|p{4mm}<{\centering}@{}} $\mathcal{D}_1$ & $\mathcal{B}_1'$ \end{tabular} 
                     & \begin{tabular}[c]{@{}p{4mm}<{\centering}|p{4mm}<{\centering}@{}} $\mathcal{D}_1$ & $\mathcal{B}_2'$ \end{tabular} 
                     & \begin{tabular}[c]{@{}p{4mm}<{\centering}|p{4mm}<{\centering}@{}} $\mathcal{D}_2$ & $\mathcal{B}_3'$ \end{tabular}
                     & \begin{tabular}[c]{@{}p{4mm}<{\centering}|p{4mm}<{\centering}@{}} $\mathcal{D}_2$ & $\mathcal{B}_4'$ \end{tabular} \\ \hline
                CUB  & \begin{tabular}[c]{@{}p{4mm}<{\centering}|p{4mm}<{\centering}@{}} $\mathcal{B}_1$ & $\mathcal{D}_1'$ \end{tabular} 
                     & \begin{tabular}[c]{@{}p{4mm}<{\centering}|p{4mm}<{\centering}@{}} $\mathcal{B}_1$ & $\mathcal{D}_2'$ \end{tabular} 
                     & \begin{tabular}[c]{@{}p{4mm}<{\centering}|p{4mm}<{\centering}@{}} $\mathcal{B}_2$ & $\mathcal{D}_3'$ \end{tabular}
                     & \begin{tabular}[c]{@{}p{4mm}<{\centering}|p{4mm}<{\centering}@{}} $\mathcal{B}_2$ & $\mathcal{D}_4'$ \end{tabular} \\ \hline
                Aircraft & \begin{tabular}[c]{@{}p{4mm}<{\centering}|p{4mm}<{\centering}@{}} $\mathcal{A}_1$ & $\mathcal{C}_1'$ \end{tabular} 
                        & \begin{tabular}[c]{@{}p{4mm}<{\centering}|p{4mm}<{\centering}@{}} $\mathcal{A}_1$ & $\mathcal{C}_2'$ \end{tabular} 
                        & \begin{tabular}[c]{@{}p{4mm}<{\centering}|p{4mm}<{\centering}@{}} $\mathcal{A}_2$ & $\mathcal{C}_3'$ \end{tabular}
                        & \begin{tabular}[c]{@{}p{4mm}<{\centering}|p{4mm}<{\centering}@{}} $\mathcal{A}_2$ & $\mathcal{C}_4'$ \end{tabular} \\ \hline
                Cars & \begin{tabular}[c]{@{}p{4mm}<{\centering}|p{4mm}<{\centering}@{}} $\mathcal{C}_1$ & $\mathcal{A}_1'$ \end{tabular} 
                     & \begin{tabular}[c]{@{}p{4mm}<{\centering}|p{4mm}<{\centering}@{}} $\mathcal{C}_1$ & $\mathcal{A}_2'$ \end{tabular} 
                     & \begin{tabular}[c]{@{}p{4mm}<{\centering}|p{4mm}<{\centering}@{}} $\mathcal{C}_2$ & $\mathcal{A}_3'$ \end{tabular}
                     & \begin{tabular}[c]{@{}p{4mm}<{\centering}|p{4mm}<{\centering}@{}} $\mathcal{C}_2$ & $\mathcal{A}_4'$ \end{tabular} \\ \hline
                \end{tabular}
                \end{center}
                \caption{The source nets for network customization of category recognition on four fine-grained 
                datasets, whose name is abbreviated as ``Dogs'', ``CUB'', ``Aircraft'' and ``Cars'', respectively.}
                \label{table:srcnet-category}
                \end{table}

                \begin{table*}[]
                \begin{center}
                \begin{threeparttable}
                \begin{tabular}{c|p{32mm}<{\centering}|p{32mm}<{\centering}||p{32mm}<{\centering}|p{32mm}<{\centering}}
                \hline
                \multirow{2}{*}{\textbf{Model}} & \multicolumn{4}{c}{\textbf{Category Sets}}\\ \cline{2-5}
                 & Stanford Dogs $\mathcal{D}_1$ & Stanford Dogs $\mathcal{D}_2$ & FGVC-Aircraft $\mathcal{A}_1$ & FGVC-Aircraft $\mathcal{A}_2$ \\ \hline
                Source Net & \begin{tabular}[c]{@{}c@{}}$\mathcal{S}_1^\text{\tiny dog}$(87.4), $\mathcal{S}_2^\text{\tiny dog}$(87.3)\end{tabular} & 
                \begin{tabular}[c]{@{}c@{}}$\mathcal{S}_3^\text{\tiny dog}$(87.9), $\mathcal{S}_4^\text{\tiny dog}$(87.7) \end{tabular} & 
                \begin{tabular}[c]{@{}c@{}} $\mathcal{S}_1^\text{\tiny air}$(70.1), $\mathcal{S}_2^\text{\tiny air}$(71.3)\end{tabular} & 
                \begin{tabular}[c]{@{}c@{}} $\mathcal{S}_3^\text{\tiny air}$(65.4), $\mathcal{S}_4^\text{\tiny air}$(65.2) \end{tabular} \\ \hline
                Component Net & $88.2^{{\bf \green{\uparrow 0.8, 0.9}}}$ & $88.5^{{\bf \green{\uparrow 0.6, 0.8}}}$ & 
                $71.5^{{\bf \green{\uparrow 1.4, 0.2}}}$ & $66.4^{{\bf \green{\uparrow 1.0, 1.2}}}$ \\ \hline
                Target Net & $88.4^{{\bf \green{\uparrow 1.0, 1.1}}}$ & $88.6^{{\bf \green{\uparrow 0.7, 0.9}}}$ & 
                $71.8^{{\bf \green{\uparrow 1.7,0.5}}}$ & $66.8^{{\bf \green{\uparrow 1.4,1.6}}}$ \\ \hline
                \end{tabular}
                \vspace{1mm}
                \begin{tabular}{c|p{32mm}<{\centering}|p{32mm}<{\centering}||p{32mm}<{\centering}|p{32mm}<{\centering}}
                \hline
                \multirow{2}{*}{\textbf{Model}} & \multicolumn{4}{c}{\textbf{Category Sets}}\\ \cline{2-5}
                 & CUB-200-2011 $\mathcal{B}_1$ & CUB-200-2011 $\mathcal{B}_2$ & Cars $\mathcal{C}_1$ & Cars $\mathcal{C}_2$ \\ \hline
                Source Net & \begin{tabular}[c]{@{}c@{}}$\mathcal{S}_1^\text{\tiny bird}$(74.5), $\mathcal{S}_2^\text{\tiny bird}$(74.8)\end{tabular} & 
                \begin{tabular}[c]{@{}c@{}}$\mathcal{S}_3^\text{\tiny bird}$(73.9), $\mathcal{S}_4^\text{\tiny bird}$(74.0) \end{tabular} & 
                \begin{tabular}[c]{@{}c@{}} $\mathcal{S}_1^\text{\tiny car}$(69.5), $\mathcal{S}_2^\text{\tiny car}$(71.1)\end{tabular} & 
                \begin{tabular}[c]{@{}c@{}} $\mathcal{S}_3^\text{\tiny car}$(71.2), $\mathcal{S}_4^\text{\tiny car}$(71.3) \end{tabular} \\ \hline
                Component Net & $75.4^{\bf \green{\uparrow 0.9,0.6}}$ & $74.8^{\bf \green{\uparrow 0.9, 0.8}}$ & 
                $72.1^{\bf \green{\uparrow 2.6,1.0}}$ & $72.8^{\bf \green{\uparrow 1.6,1.5}}$ \\ \hline
                Target Net & $75.8^{\bf \green{\uparrow 1.3,1.0}}$ & $75.4^{\bf \green{\uparrow 1.5,1.4}}$ & 
                $72.5^{\bf \green{\uparrow 3.0,1.4}}$ & $73.1^{\bf \green{\uparrow 1.9,1.8}}$ \\ \hline
                \end{tabular}

                \begin{tablenotes}
                \item[\green{$\uparrow$}] denotes performance improvement compared with the corresponding source network.
                \end{tablenotes}
                \end{threeparttable}
                \end{center}
                \caption{The performance (\%) of knowledge amalgamation from source nets to component net and
                from component net to target net on four fine-grained datasets.
                }
                \label{table:category}
                \vspace{-3mm}
                \end{table*}
                
            }
        }

        \subsubsection{Learning from Varying Numbers of Teachers}
        {
            \begin{figure}
                \centering
                \includegraphics[width=0.94\linewidth]{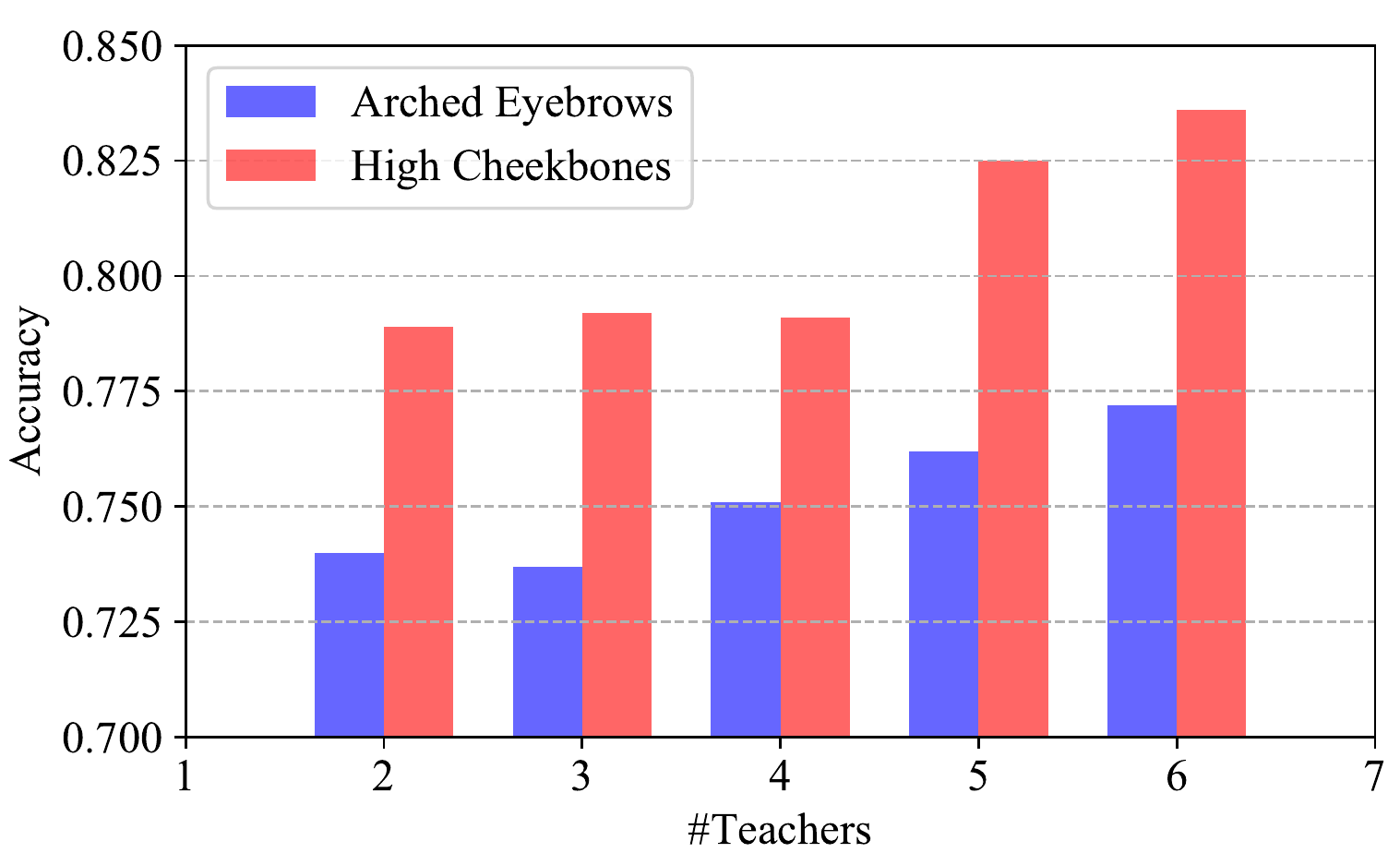}
                \caption{The performance of knowledge amalgamation for different number of source nets on the CelebA dataset. }
                \label{fig:multi-teacher}
                \vspace{-3mm}
            \end{figure}

            To investigate the effect of knowledge amalgamation for more teachers, we also conduct experiments in which
            varying numbers of source nets are amalgamated into a single component net.
            The experiments are implemented on two face attribute recognition tasks, 
            ``arched eyebrows'' and ``high cheekbones'', 
            as shown in Fig.~\ref{fig:multi-teacher}.
            With more teachers, the performance tends to be better for both face attribute recognition tasks.
            By integrating more teachers, the student network may potentially ``absorb'' more complementary knowledge from multiple teachers 
            and significantly reduce erroneous guidances from teachers.

        }

        \subsubsection{Ablation Study}
        {
            Ablation study is conducted on several attributes to investigate the effectiveness of the modules adopted in our 
            proposed approach.
            Specifically, we verify the effectiveness of each module by comparing the whole model to the model without the 
            corresponding module.
            Additional compared method is knowledge distillation, which does not contain transfer bridge module and teacher
            selective learning strategy. 

            The results shown in Tab.~\ref{table:ablation_study} demonstrate that both transfer bridge and selective 
            learning strategy significantly improve the performance of the model.
            The transfer bridges deliver the partial task-demanded intermediate features of teacher networks to the student 
            network, which provide more supervision to the student network compared to knowledge distillation.
            And the selective learning strategy takes the most confident teacher as the learning target, which can significantly
            reduce the misleading information provided by teachers.

            {
                \begin{table}[]
                \begin{center}
                \begin{tabular}{|c|p{12mm}<{\centering}|p{22mm}<{\centering}|p{12mm}<{\centering}|}
                \hline
                \multirow{2}{*}{\textbf{Module}} & \multicolumn{3}{c|}{\textbf{Attributes}} \\ \cline{2-4} 
                 & black hair & mouth slightly open  & brown hair \\ \hline
                KD~\cite{hinton2015distilling} & 87.1 & 90.2 & 87.4 \\ \hline
                wo/TB & 87.4 & 91.3 & 87.7 \\ \hline
                wo/TS & 87.4 & 91.0 & 87.8 \\ \hline
                whole model & \textbf{87.8} & \textbf{91.4} & \textbf{88.0} \\ \hline
                \end{tabular}
                \end{center}
                \caption{The performance (\%) for ablation study on the CelebA dataset.  
                \emph{KD} denotes knowledge distillation (baseline).
                \emph{TB} denotes transfer bridge.
                \emph{TS} denotes teacher-selective learning.}
                \label{table:ablation_study}
                \vspace{-4mm}
                \end{table}
            }
        }

        \subsubsection{{One-shot Amalgamation}}
        {
            To further explore network customization methods, we compare an intuitive variant of
            our proposed dual-stage method: one-shot amalgamation.
            In this scenario, multiple sources nets are directly amalgamated into target net without the
            component net as the intermediate byproduct.
            The experiments are conducted on two face attribute recognition tasks, as shown in Tab.~\ref{table:one-shot-amalga}.
            The results demonstrate that two-stage amalgamation method outperforms the 
            one-shot one on both of face attributes.
            Because one-shot amalgamation is required to simultaneously learn knowledge from more source networks, 
            instead of learning from few component nets adopted in two-stage method,
            it potentially complicates the optimization of student net and leads to poorer performance.

            {
                \begin{table}[]
                \begin{center}
                \begin{tabular}{|c|c|c|c|}
                \hline
                \multirow{2}{*}{\textbf{Method}} & \multicolumn{2}{c|}{\textbf{Attributes}} \\ \cline{2-3} 
                 & Black Hair & Blond Hair \\ \hline
                one-shot amalgamation & 85.6 & 86.1 \\ \hline
                two-stage amalgamation & \textbf{87.6} & \textbf{95.1} \\ \hline
                \end{tabular}
                \end{center}
                \caption{The performance (\%) comparison between one-shot amalgamation and two-stage amalgamation on the CelebA dataset.}
                \label{table:one-shot-amalga}
                \vspace{-4mm}
                \end{table}
            }

        }
    }
}

\section{{Conclusion and Future Work}}
{
    In this paper, we propose an adaptive knowledge amalgamation method
    to learn a user-customized student network, 
    without accessing human annotations,
    from a pool of single- or multi-task teachers working on
    distinct tasks. 
    This is achieved specifically via a 
    dedicated dual-stage approach.
    In the first stage,
    source nets covering the same task
    are clustered into groups, from each of which
    a component net is learned; in the second,
    the components are further amalgamated into the
    user-specified target net. 
    Both stages undergo a similar knowledge amalgamation
    process, where for each unlabelled sample, 
    the student learns the features and predictions
    of only one teacher, taken to be the one with the least
    prediction ambiguity. The feature learning is achieved 
    via a dedicated transfer bridge,
    in which the features of the student are aligned with
    those of the selected teacher for learning. 
    We conduct experiments on several datasets
    and show that, the learned student that comes in 
    a compact size, 
    yields consistent superior results to those 
    of the teachers in their own specializations.
    For future work, we plan to customize networks using teachers of different network architectures.
}

\noindent {\bf Acknowledgments.} This work is supported by  National Key Research and Development Program (2016YFB1200203) , National Natural Science Foundation of China (61572428,U1509206),  Key Research and Development Program of Zhejiang Province (2018C01004), and the Program of International Science and Technology Cooperation (2013DFG12840).

{\small
\bibliographystyle{ieee_fullname}
\bibliography{mybib}
}

\end{document}